\pdfoutput=1

\documentclass[11pt]{article}

\usepackage[]{ACL2023}

\usepackage{times}
\usepackage{latexsym}
\usepackage{pgfplots}
\usepackage{pgfplotstable}
\pgfplotsset{compat=1.7}
\usepackage{tikz}
\usepackage{amsfonts}
\usepackage{colortbl}

\usepackage[T1]{fontenc}

\usepackage[utf8]{inputenc}

\usepackage{microtype}

\usepackage{inconsolata}

\usepackage{todonotes}

\newcommand{\qa}{$(q_i,a_i)$~}

\usepackage[inline]{enumitem}
\usepackage{multirow}

\title{QUADRo: Dataset and Models for QUestion-Answer Database Retrieval}

\author{Stefano Campese \\
  University of Trento \\
  Amazon Alexa AI \\
  \texttt{stefano.campese@unitn.it} \\\And
  Ivano Lauriola \\
  Amazon Alexa AI \\
  \texttt{lauivano@amazon.com} \\\And
  Alessandro Moschitti \\
  Amazon Alexa AI \\
  \texttt{amosch@amazon.com} \\
  }

\begin{document}
\maketitle
\begin{abstract}
An effective paradigm for building Automated Question Answering systems is the re-use of previously answered questions, e.g., for FAQs or forum applications. Given  a database (DB) of question/answer (q/a) pairs, it is possible to answer a target question by scanning the DB for similar questions. 
In this paper, we scale this approach to open domain, making it competitive with other standard methods, e.g., unstructured document or graph based. For this purpose, we (i) build a large scale DB of $\approx 6.3$M q/a pairs, using public questions, (ii) design a new system based on neural IR and a q/a pair reranker, and (iii) construct training and test data to perform comparative experiments with our models.
 We demonstrate that Transformer-based models using (q,a) pairs outperform models only based on question representation, for both neural search and reranking. Additionally, we show that our DB-based approach is competitive with Web-based methods, i.e., a QA system built on top the BING search engine, demonstrating the challenge of finding relevant information.
Finally, we make our data and models available for future research. 

\end{abstract}

\section{Introduction}

In recent time two main QA paradigms have gain more and more attention from both the industrial and research community: graph- and web-based, where the web in academic research often refers to some large repository of text, e.g., Wikipedia~\cite{yang2015wikiqa, nguyen2016ms, thorne-etal-2018-fever}.
One approach neglected from the latest work is QA based on q/a pair DB: early work on QA for forum or FAQ~\cite{nakov-etal-2016-semeval-2016, Shen_Rong_Jiang_Peng_Tang_Xiong_2017, 10.1145/2838931.2838934} have pointed out that when answers are available in the DB together with the questions, the resulting system can be very accurate. The main challenge to their general usage is the typical specificity of the DB, also associated with a limited availability of q/a pairs. 

However, since early works, QA applications and related datasets have multiplied, and nowadays a large set of pairs is available, which may represent general user information need. The main challenge is then the design of an effective question retrieval that can find semantically similar questions, even in case the surface form and syntax of questions is rather different. This effective retrieval would   increase the generality of the resulting QA system.

%
%
%
In this work, we take a step forward by scaling up the standard QA systems based on q/a DBs, generalizing them to open domain QA. 
%
For this purpose, we introduce an end-to-end pipeline, here named QUestion-Aswer Database Retrieval (QUADRo). The system comprises three main components: (i) a large-scale DB of questions and their correct answers obtained as mixture of various unlabeled and labeled open-domain QA sources, including Quora~\cite{quoraqp}, GooAQ~\cite{gooaq2021}, and WikiAnswer~\cite{Fader14}, ($\approx$6.3M q/a pairs); (ii) an efficient neural search engine for q/a pairs retrieval, based on the latest neural IR research~\cite{karpukhin2020dense}; and (iii) an effective answer selector, which exploits both, question similarity and question/answer relevance. 
%
Given an input question, the system queries the internal database through the neural search engine looking for semantically equivalent questions. 
Then, the answer sentence selector model is applied to the output of the search engine to select the most similar question. Finally, the answer in the q/a pair that obtained the highest score from the selector is returned to the user. 
%

%
%

To design and evaluate our proposed end-to-end QA pipeline, we built two new resources which consists of a large q/a pair DB and a dataset annotated in a ranking fashion to evaluate retrieval and answer selection models. 
The resource consists of 15,211 open-domain input questions. For each input question, we annotated the top 30 q/a pairs retrieved in terms of question similarity, and we marked them as semantically equivalent or not (binary label) with respect to the input. 
We select candidates from the DB using a Transformer model trained for question-question similarity. 
%
Differently from existing resources, our dataset is annotated with respect to both, question-question similarity and question-answer correctness. 

Our experiments shows: (i) the high quality of our datasets and its suitability for advancing the state of the art in DB-based QA, and (ii) our system produced competitive results in terms of accuracy compared to popular open-domain web-based QA pipelines. 
We will make our annotated dataset, the collected database of q/a pairs, and our models for search engine and reranking available to the research community.

\section{Related Work}
Our work is related to question similarity task, q/a DB retrieval, and construction of q/a DB/datasets. 
\paragraph{Question Similarity} Previous work relevant to our study is about Duplicate Question Detection (DQD), which
%
is a well known problem, under the umbrella of Semantic-Textual-Similarity (STS) tasks. The latter aim at identifying when two questions are semantically equivalent or not. Several methods have been proposed over the years to solve this problem. 
Early approaches focused on the extraction and creation of several types of lexical~\cite{cai2011learning}, syntactic~\cite{moschitti2006efficient} and heuristic~\cite{filice2017kelp} features to measure the similarity between two questions. 
Lately, translation- and topic-based modeling approaches have been explored, e.g.,~\cite{wu2016ecnu}.

DQD received a huge boost with the advent of embedding representations such as Word2Vec and Glove, e.g.,
\cite{charlet2017simbow} 
%
%
and ELMO~\cite{https://doi.org/10.48550/arxiv.1802.05365}, used by~\citet{fadel2019tha3aroon}  to compute sentence-level embeddings for the two individual questions. 

More recently pretrained Transformers set the state of the art for STS and DQD~\cite{chandrasekaran2021evolution}.
\citet{Peinelt2020tBERTTM} proposed tBERT, a Transformer model that takes the concatenation of the two questions as input, providing a joint and contextualized representation of the pair.

\paragraph{q/a DB retrieval}


QA based on q/a DB is a known paradigm, which relies on (i) a curated database (or collection) of questions and their answers and (ii) a model to query the database for finding an equivalent question, and then returning the associated answer to the user.
%
These systems are typically used in FAQ, chatbot, and community QA.

Various DB QA systems have been proposed in the literature.
%
%
%
\citet{othman2019enhancing} introduced WEKOS, a system able to identify semantically equivalent questions from a FAQ database. 
%
The model, based on k-means clustering and heuristics, was tested on a dataset based on Yahoo Answer, showing impressive performance compared to the existing systems. 
\citet{mass-etal-2020-unsupervised} proposed a new ensemble system which combines BM25 with two BERT~\cite{devlin-etal-2019-bert} models. 
The same authors also explored systems based on a GPT2 model to generate question and/or answer paraphrases to fine-tune the BERT models on low resources FAQ datasets with promising results. 
Similarly,~\citet{sakata2019faq} proposed a method based on BERT and TSUBAKI, an efficient retrieval architecture based on BM25~\cite{shinzato2012tsubaki}, to retrieve from a database of FAQ the most similar questions looking at query-question similarity. 

Our approach introduces two novelties with respect to the methods above: (i) it uses training data, which makes our system more accurate, and (ii) we target very large DB (6M items), and aim at scaling our QA system to open domain. Previous work has just targeted specific FAQ tasks. 

\paragraph{Datasets and resources}
\label{sec:related-res}
Various resources have been made available to train models for DQD and DB-based QA.
One of the most popular resources is the QuoraQP dataset.
This consists of 404,290 pairs of questions extracted from the Quora website, and annotated as having the same meaning or not. 
An extension of the original dataset was released by~\citet{wang2020match2}, which
 consists of the original question pairs concatenated with answers to the second questions, extracted from the original Quora threads. 
%
%
Another popular resource is the CQADupStack~\cite{10.1145/2838931.2838934} dataset,  originally released as a benchmark for Multi-Domain Community QA. It consists of questions coming from different threads sampled from twelve StackExchange subforums. The dataset contains annotations for similar and related questions. 
For a small portion of question pairs, there is an annotated answer. On average the dataset contains $\approx5.03\%$ of duplicated questions. 

WikiAnswers is another well know resource, which was
 built by clustering together questions classified as paraphrases by WikiAnswers users. 
The dataset consists on 30,370,994 clusters containing an average of 25 questions per cluster. Unfortunately, the answers are large paragraphs, which are not suitable for DQD.

SemEval-2016 Task 3 challenge~\cite{nakov-etal-2016-semeval-2016} introduced another famous resource for DQD in the cQA domain, where Question-Comment, Question-External Comment, and Question-Question Similarity, are annotated. 
Although this dataset can be used into a reranking settings, the amount of queries is limited. Moreover, all the data is extracted from specific domains. 

In general, the resources above have some limitations:
(i) most datasets do not include answers, (ii) there is no guarantees on the quality, and (iii) with the exception of SemEval, these resources are based on pairs of questions (e.g.: QuoraQP) rather than question ranking, preventing the possibility to study search and ranking problems.

Our DB and dataset instead enables research on large-scale retrieval of semantically equivalent questions, associated with correct answers, which are essential to build large-scale DB-based QA.


\section{QUADRo}
\label{models}

\begin{figure}[t]
\centering
  \includegraphics[width=\linewidth]{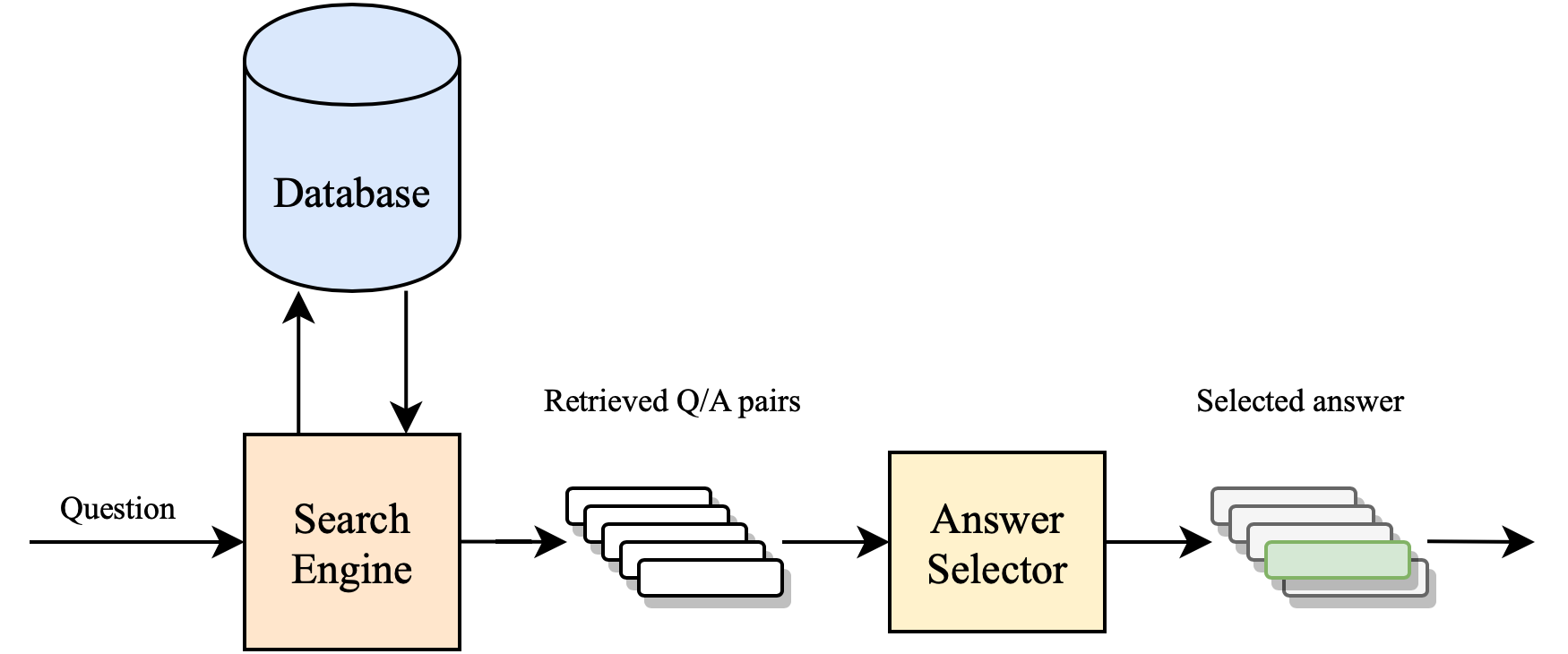}
  \caption{QUADRo workflow. First, a search engine queries the database of known q/a pairs. Then, a sentence selector finds an equivalent question, whose answer is returned to the user.}
  \label{fig:qrs_design}
\end{figure}

QUestion Answer Database Retrieval (QUADRo) is an end-to-end framework for open domain QA based on q/a DB.
%
%
As described in \figurename~\ref{fig:qrs_design}, QUADRo consists of three main components: a large-scale database of questions and correct answers, an efficient search engine to query the database, and an answer selector. 
Given an input question, QUADRo queries the database looking for semantically equivalent (or similar) questions and returns the top similar q/a pairs. Then, an answer sentence selector model reranks the pairs and returns the answer associated with the top 1 pair to the user.
The individual modules of the system are described in the following sections.

\paragraph{Search Engine}


This is based on recent findings in neural retrieval, including Dense Passage Retrieval~\cite{karpukhin2020dense}. 
It consists of a siamese Bi-Encoder Transformer~\cite{reimers-2019-sentence-bert} (also known as Sentence-Transformer) network. 
The first branch encodes the input question $t$ as \emph{[CLS] $t$ [EOS]}, whereas the second branch encodes question/answer pairs \qa from the database as \emph{[CLS] $q_i$ [SEP] $a_i$ [EOS]}. 
The cosine similarity between the representations extracted from the two branches expresses the level of the semantic similarity between the question and the q/a pair. 
Let $\delta : \Sigma* \rightarrow \mathbb{R}^d$ be the Transformer function which maps a text $s\in\Sigma*$ (either a question $q_i\in\mathcal{Q}$, an answer $a_i\in\mathcal{A}$, or a pair $(q_i,a_i)\in\mathcal{Q}\times\mathcal{A}$) into a $d$-dimensional embedding. 
The final score assigned to a target question $t\in\mathcal{Q}$ and an element of the database $(q_i,a_i)$ is $\frac{\delta(t)^\top\delta(q_i,a_i)}{\|\delta(t)\| \|\delta(q_i,a_i)\|}$, where $\|\cdot\|$ denotes the 2-norm of a vector.
When the user asks a new question, the bi-encoder (i) encodes the input query, (ii) applies cosine distance between the resulting embedding and all q/a pairs, and (iii) returns the most similar $k$  pairs. Note that the computation of the embeddings for each stored pair can be executed offline. 
\paragraph{Answer Selector}
After the retrieval stage, an answer selector (or reranker) model re-ranks the pairs returned by the search engine and selects the final answer to be served to the user. 
Formally, let $\mathcal{R} = \{(q_i,a_i)\}_{i=1}^k$ be the set of $k$ returned pairs for a given target question $t$. 
The answer selector $r: \mathcal{Q} \times \mathcal{R} \rightarrow \mathbb{R}$ assigns a score to each  triplet, $r(t,q_i,a_i)$, and returns the answer associated with highest ranked pair, i.e., $\arg\max_{i=1\dots k} r(t,q_i,a_i)$.
%
Inspired by the Contextual Sentence Selector framework~\citep{lauriola2021answer}, which is the state of the art for open-domain answer selection, we use a Transformer model to encode and rank triplets. 
The input of the transformer is encoded as 
%
%
%
\emph{[CLS] $t$ [SEP] $a_i$ [SEP] $q_i$ [EOS]}.
Note that this allows to jointly modeling the semantic dependencies between the two questions, e.g., their similarity, and the relevance of the answer to both questions. 

\section{Database and Dataset Construction}
As Section~\ref{sec:related-res} highlights, existing resources do not enable the development of an open domain q/a based system. There are several corpora available but a dataset for training and testing retreival and ranking from a large DB is missing.
In this section, we describe (i) our collected q/a pair DB, and (ii) the design choices, the sampling strategy, and the annotation procedure of the associated dataset. 

\subsection{The Database}
The DB of QUADRo consists of questions and their answers collected from various heterogeneous public high-quality annotated open-domain QA datasets, including: 
%
%
GooAQ, WQA~\cite{zhang-etal-2021-joint}, WikiAnswer, CovidQA~\cite{moller2020covid}, and HotpotQA~\cite{yang2018hotpotqa}. 
We extracted questions and the associated correct answer text span from these datasets, and ingested these q/a pairs into our DB. It should be noted that we do not require the answer to be necessarily correct, as our final model, also operates an answer selection step, which can choose the answer with the highest probability to be correct. 

Beyond annotated resources, we enhanced our database with various sets of artificially generated q/a pairs.
First, we considered question from QuoraQP. Differently from the other datasets above, QuoraQP is designed for question duplication detection task and not QA.
Thus it simply consists of questions pairs. 
Some answers are available for a small fraction of questions, which were selected through a heuristic approach based on the rank of users' content in Quora threads~\cite{wang2020match2}. 
To expand the QuoraQP, we collected the missing answers using as a similar approach described by~\citet{zhang-etal-2021-joint}: given an input question, we queried a 2020 CommonCrawl snapshot\footnote{https://commoncrawl.org/2020/?utm\_sou} using BM25 and we selected the 200 most relevant documents. Then, documents are split into sentences and a state-of-the-art sentence selector~\cite{lauriola2021answer} is used to select the top-ranked passage as the answer.
In addition, the score returned by the passage reranker model can be effectively used to measure the likelihood of the answer to be correct. We applied a threshold to this score and accepted only the top 10\% of q/a pairs. This guarantees a higher level of answer quality in our database.
We manually labelled 200 randomly selected answers from this unsupervised set as quality assessment, observing an accuracy of 93.0\%. 
Finally, we also ingested q/a pairs from ELI5~\cite{fan2019eli5}. 
This dataset consists of questions collected from three subreddits for which the answers have been ranked by the users' up-votes in the thread. 
Although this heuristic removed part of the noise of the dataset, to ensure the maximum quality we keep only the 50\% of the q/a pairs with the highest sentence selector score. After a manual annotation, we estimate the 84.3\% of accuracy. Our final DB contains $\approx6.3$ millions of English questions and their correct answers pairs. Further details and statistics are reported in Table~\ref{tab:qrs_resources}.
We manually estimated a correctness of $\approx 92.6$\% of the answer with respect to stored questions.


\begin{table}[t]
\resizebox{\linewidth}{!}{
\begin{tabular}{lllll}
    \hline
    \textbf{Source} & \textbf{QA} & \textbf{Q} & \textbf{Q lenght} & \textbf{A lenght} \\
    \hline
    \multicolumn{5}{c}{\textbf{Labeled}}\\
    \hline
        GooAQ & 3.1M & 2.9M & $9.1_{\pm2.3}$ & $45.9_{\pm18.9}$\\
        WQA & 391K & 80.5K & $7.5_{\pm3.2}$ & $24.8_{\pm11.3}$\\
        WikiAnswer & 2.3M & 2.3M & $9.1_{\pm2.5}$ & $60.3_{\pm117.3}$ \\
        CovidQA    & 2K & 1.9K & $10.6_{\pm4.1}$ & $15.8_{\pm17.1}$\\

        HotpotQA    & 64K & 64K & $20.4_{\pm10.6}$ & $4.1_{\pm2.4}$\\
    \hline
    \multicolumn{5}{c}{\textbf{Unlabeled}}\\
    \hline
        Quora Match & 230K & 170K & $12.5_{\pm6.7}$ & $38.8_{\pm20.5}$ \\
        QuoraQP & 219K & 134K & $9.6_{\pm2.9}$ & $25.1_{\pm10.8}$ \\
        ELI5    & 58.9K & 58.7K & $17.6_{\pm9.10}$ & $60.7_{\pm28.1}$\\
    \hline
    \textbf{Total}    & \textbf{6.3M} & \textbf{5.7M}\\ 
    \hline
\end{tabular}}
\caption{Main statistics of QUADRo database, QA= q/a pairs, Q= unique questions.}
\label{tab:qrs_resources}
\end{table}

\subsection{Question Ranking dataset} However, a QA system able to exploit it, requires an additional annotation for training and evaluation. The main missing annotation is the annotation of semantic equivalence between questions. Since this is impossible to obtain for all pairs in our DB, we select a subset of questions and then retrieve and annotate to the top $k$. In order to obtain high quality data, we defined an annotation process, which also exploits the availability of an answer for each retrieved question.

\label{sec:dataset_construction}
We randomly selected 15,211 questions from our DB, and removed them from it to avoid bias for the next experiments. 
For each of the questions above, we ran the search engine of QUADRo\footnote{We describe the configuration of the search engine used to collect q/a pairs in Section~\ref{sec:exps}.} and retrieved 30 most similar questions and their answers. 
We used Amazon Mechanical Turk (AMT) to annotate triplets $(t,q_i,a_i)$, target question, retrieved question, and answer as correct/incorrect, where this binary label represents the semantic equivalence or not, respectively, between $t$ and $q_i$. 
%


%

Given the complexity and subjectivity of the task, we insured high quality of our annotation, applying several procedures, some of which are completely novel: 
First, we provide the following definition of equivalence to the annotators: two questions are equivalent iff they (i) have the same meaning AND (ii) share the same answers. 
While expressing the same meaning is theoretically enough to assess semantic equivalence of the questions, in practice, there are several sources of ambiguity, e.g., it is well known that the meaning depends on the context, and the latter may be underspecified. The answer instead provides a strong context that can focus the interpretation of the annotator on the same meaning both questions\footnote{A few examples are reported in the appendix.}.  

One source of complexity was given by the fact that we do not have 100\% accurate answers, thus, we clarified with the annotators that they should use answers to help their judgment, but these are not necessarily correct. 
We ran a pilot annotation batch to evaluate the performance of the annotators and to quantify the benefits of including the answer into the annotation interface. 
Our manual evaluation on 200 annotations showed that adding the answer in the annotation workflow reduces the relative error rate for this task up to 45\%, for an absolute accuracy of 94\%. 
In other words, the answer is a precious source of information which, beyond the modeling part where previous work already showed the importance of the q/a relevance, can significantly help the human judgment and thus the quality of the dataset. 
To our knowledge, our is the first resource for question-question similarity  exploiting answer as context in order to perform accurate annotation.
Second, we used two distinct annotators, and a third one to resolve the tie cases, thus the final label is computed with the majority vote.
Third, we limited the annotation to turkers with documented historical annotation activities, including master turkers with at least 95\% approval rate and 100 approved tasks.

Finally, we introduced a set of positive and negative control triplets that we used to filter bad annotations. 
%
Each annotation task consists of 7 triplets of which 2 control triplets, one negative and one positive.
Positive control triplets were designed to be as clear as possible and sufficiently simple. Differently, negative triplets are automatically defined by selecting two random questions. 
Answering incorrectly to at least one of the control pairs caused the rejection of the annotation task. In addition, according to the control pair, if the same turker failed more than 10\% of the total assigned HITs, all their HITs are discarded and the turker is blacklisted, precluding the execution of further annotations. Guidelines, some examples of control triplets, and further details are showed in the Appendix~\ref{appendinx:annotation}.

\section{Experiments}\label{sec:exps}

We ran various distinct sets of experiments to assess the quality of our dataset and demonstrate the competitiveness of our QA approach.
First, we analyze the retrieval and ranking tasks thanks to our new dataset.
Then, we fine-tune and evaluate QUADRo components (both retrieval and rankers) trained on our dataset. Finally, we compare our approach against strong Web-based open-domain QA systems.

\begin{table}[t]
\small
\centering

\begin{tabular}{lccc}
    \hline
     \textbf{Split} & \textbf{Q} & \textbf{QA} & \textbf{pos/neg QA} \\
    \hline
        Train  & 11711 & $28.9_{\pm10.3}$ & $6.1_{\pm7.9}$ / $22.7_{\pm11.5}$\\
        Dev. & 1500 &  $30_{\pm0}$ & $4.8_{\pm5.3}$ / $25.2_{\pm5.3}$\\
        Test & 2000 & $30_{\pm0}$ & $4.7_{\pm5.1}$ / $25.3_{\pm5.1}$ \\
    \hline    
\end{tabular}
\caption{Data splits. Q= num. queries, QA= q/a pairs per query.}
\label{tab:qrc}
\end{table}

\subsection{Datasets}
We ran most of our experiments on our collected dataset for question ranking. 
We divided our dataset into training, development, and test questions as described in Table~\ref{tab:qrc}. 

Additionally, we ran end-to-end QA evaluations on questions sampled from various open-domain sources, including QuoraQP and:

\textbf{Natural Questions}: NQ~\cite{kwiatkowski2019natural} is a popular open-domain QA dataset which consists of questions sampled from Google traffic. A wikipedia page containing a long and a short answer is associated with each question. Consequently, each question is virtually answerable by a web based QA system.


\textbf{TriviaQA}: TriviaQA~\cite{joshi2017triviaqa} is a QA dataset containing over 95,000 open-domain q/a pairs authored by trivia enthusiasts and independently gathered evidence documents. Trivia questions are designed to be more challenging, complex, and compositional compared to the other datasets.

\subsection{Model Training}
\label{training}
We used our dataset to train and  evaluate both, retrieval and ranking models. 

\paragraph{Retrieval Models:} we built the bi-encoder (described in Section~\ref{models}) with RoBERTa continuously pre-trained on $\approx 180$M semantic text similarity pairs\footnote{See appendix~\ref{appendix:pretraining} for further pre-training details.}. 
We considered the following two input configurations:
QQ: the model encodes the input question, $t$, and the evaluation questions $q_i$ in the first and second transformer branches (with shared weights). 
QQA: the model encodes $t$ in the first branch, and the concatenation of the $q_i$ and $a_i$ in the second one.
After this procedure, the model computes the cosine similarity between the two generated representations. 
We first fine-tuned the models on QuoraQP with artificially generated answers as previously described. Then, we fine-tuned the resulting model on our ranking dataset.

\paragraph{Reranking Models:} we start from the state-of-the-art sentence selector model proposed by~\citet{lauriola2021answer}. The model consists of an Electra-base trained on ASNQ on triplets of the form \emph{[CLS] question [SEP] answer [SEP] context [EOS]}, where \emph{context} refers to additional sentences which are relevant for the answer.
The checkpoint was then fine-tuned on QuoraQP and lately on our dataset. As in our case the $a_i$ can be considered as the context of both $t$ or $q_i$, or alternatively, $q_i$ may be considered the context of $t$ and $a_i$. Therefore, the use of a pre-trained model with context is promising.
Similarly to the retrieval training, we considered various configurations, including (i) QQ only using  $(t,q_i)$ pairs; (ii) QAQ corresponding to $(t,a_i,q_i)$, where the question acts as a context for the answer, (iii) QQA corresponding to $(t,q_i,a_i)$, where $a_i$ is the context, and (iv) QA encoding $(t,a_i)$, i.e., a standard sentence selector for QA.
For both, retrieval and reranker the model score ranks the q/a pairs with respect to each input question. 

\subsection{Metrics} 
We measure the performance of  QA systems with Accuracy in providing correct answers, i.e., the percentage of correct responses, which also refers to Precision-at-1 (P@1) in the context of reranking. 
We also use standard metrics for ranking: Mean Average Precision (MAP), Mean Reciprocal Rank (MRR), and Hit-rate@k. 

\begin{figure}[t]
\centering
\resizebox{0.80\linewidth}{!}{
\begin{tikzpicture}
\begin{axis}[
	xlabel=K,
	ylabel= \% HIT Rate,
    legend style={at={(.6,.2)},anchor=west},
    xtick={1,5,10,15,20,25,30},
    grid=both,
    grid style={line width=.1pt, draw=gray!10},
    major grid style={line width=.2pt,draw=gray!50},
    minor grid style={line width=.2pt,draw=gray!10},
    minor tick num=3,
    ]


\addplot[color=blue,mark=x,line width=.2mm] coordinates {
	(1, 42.60)
	(3, 58.0)
	(5, 63.55)
	(7, 66.53)
	(10, 69.21)
	(15, 71.86)
	(20, 73.61)
	(25, 74.90)
	(30, 76)
};


\addplot[color=orange,mark=x,line width=.2mm] coordinates {
	(1, 46.66)
	(3, 61.65)
	(5, 66.55)
	(7, 69.40)
	(10, 71.45)
	(15, 73.41)
	(20, 74.65)
	(25, 75.35)
	(30, 76)
};

\legend{Retrieval, +Ranker}
\end{axis}

\end{tikzpicture}}
\caption{Hit rate of the retrieval module and the end-to-end system (QUADRo).}
\label{fig:hit_rate}
\end{figure}
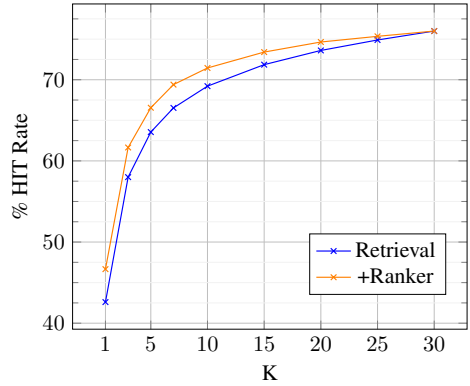

\subsection{Answer Retrieval and Ranking Performance}

As discussed in Section~\ref{sec:dataset_construction}, we used an initial retrieval model to select the q/a pairs that compose to build the question reranking dataset. 
Based on preliminar experiments conducted on existing benchmarks, we used a Sentence-RoBERTa (base) model. The model is trained on QuoraQP with QQA configuration (as described in Section~\ref{training}).
%
Similarly, we trained an initial reranker with QAQ configuration to be tested on our collected dataset.


 \figurename~\ref{fig:hit_rate} shows the hit-rates at $k$ of the sole retrieval component and the Hit-rate of the entire system, i.e., after the reranking performed by the answer selector, applied to the 30 retrieved (and annotated) answers. 

We note that:
First, as we used the retrieval mentioned above to select 30 question/answer pairs to be annotated in our dataset the plot exactly indicates the retrieval performance. Since the ranker operates on just the 30 annotated candidates, we can also evaluate its impact on an end-to-end system in an open-domain QA setting. This means that the accuracy of QUADRo (Hit@1) in answering open domain questions sampled (and removed) from the DB is 46.6\%, as indicated by the plot.

Second, as expected, the larger the $k$, the higher is the probability to retrieve a good answer. Using 30 candidates, the retrieval is able to find an answer for 76\% of the questions. The reranker boosts the retrieval performance by 4\% absolute (42.6 vs 46.6). Then, it consistently improves the retrieval. 

Note that a configuration entirely based on the answer selector (i.e. without retrieval) is infeasible due to the complexity of the model as it needs to run a Transformer for each stored pair (6.3M).
Finally, there is a considerable gap between the system accuracy (i.e.: Hit@1) and the potential accuracy (Hit@30). This suggests that our DB and datasets enable challenging future  research. Also, there is a significant amount of unanswerable queries, which open other research challenges on how learning to not answer to improve system F1.

\subsection{Retrieval and reranker evaluation}
We fine-tuned our models on the collected dataset as described in~\ref{training} and evaluated them on the test split. Details of the fine-tuning step are described in Appendix~\ref{appendix:retrieval_and_ranking_model}.
Table~\ref{tab:experimental_results} reports the performance of (i) *S-RoBERTa$_{QQA}$: the initial neural retrieval model used to select annotation data (and thus not trained on this resource), 
(ii) S-RoBERTa$_{x}$: our retrieval models fine-tuned on the collected resource, 
and (iii) Electra$_{x}$: the fine-tuned reranking models. 
As mentioned before, these models can be applied to the top 30 candidates retrieved by *S-RoBERTa$_{QQA}$, which are all annotated thus enabling their easy evaluation. 
The selection of RoBERTa for retrieval and Electra for reranking was driven by a set of preliminary experiments described in Appendix~\ref{appendix:retrieval_and_ranking_model}.


We note that the Electra (reranker) generally generally outperforms S-RoBERTa (retrieval), as it is a cross-encoder, where a single network encodes the whole triplet. Differently, S-RoBERTa uses two different encoders which provide two uncontextualized representations that are successively combined only in the final layer.
Moreover, S-Roberta using the QQA configuration highly outperforms QQ (+5.0 P@1), while this gap disappears for Electra since its cross-encoder seems enough powerful to compensate for the lack of context, i.e., the answer.

Concerning Electra models, QQA is outperformed by QQ (-0.7 P@1), mostly because the checkpoint that we used was trained by~\citet{lauriola2021answer} on QA tasks, thus the model expects the answer to be close to the query. Indeed, QAQ, which is closer to how the checkpoint was trained, improves the P@1 of QQ by 0.8. It should also be stressed the fact that 
our dataset has been annotated with respect to question-question equivalence and not question-answer correctness. Although the answers were shown to the annotators, they are biased on the question and used the answer just as support. 
Finally, we evaluated existing state-of-the-art question de-duplication models~\citep{reimers-2019-sentence-bert} consistsing on a RoBERTa cross encoder trained on QuoraQP\footnote{Public checkpoints are available \url{https://www.sbert.net/docs/pretrained_cross-encoders.html}.} (see table~\ref{tab:experimental_results}). 
Not surprisingly, our models based on the similar architectures achieve better performance thanks to the fine-tuning on the target data. We did not evaluate other existing solutions (e.g.~\citet{wang2020match2}) as models are not publicly available.

\begin{table}[t]
\resizebox{\linewidth}{!}{
 \centering
 \begin{tabular}{lccc}
 \hline
\textbf{Model} & \textbf{P@1} & \textbf{MAP} & \textbf{MRR}\\
 \hline
 
 *S-RoBERTa$_{QQA}$   & $39.1$ & $39.1$ & $50.4$  \\
 \hline
 
 S-RoBERTa$_{QQ}$   & $43.4$ & $41.6$ & $52.9$  \\
 
 S-RoBERTa$_{QQA}$ & $48.4_{\pm0.4}$ & $45.6_{\pm0.4}$ & $58.3_{\pm0.4}$ \\

  \hline
 Electra$_{QA}$ & $37.1_{\pm0.6}$ & $40.4_{\pm0.2}$ & $49.5_{\pm0.3}$ \\
 Electra$_{QQ}$ & $50.0_{\pm0.2}$ & $47.7_{\pm0.3}$ & $59.5_{\pm0.2}$ \\
 Electra$_{QQA}$ & $49.3_{\pm0.2}$ & $47.63_{\pm0.1}$ & $59.2_{\pm0.1}$ \\
 Electra$_{QAQ}$ & $\mathbf{50.8_{\pm0.2}}$ &  $\mathbf{48.4_{\pm0.1}}$ & $\mathbf{60.2_{\pm0.1}}$ \\
\hline
QP-RoBERTa$_{base}$ & ${43.5}$ & ${41.8}$ & ${54.4}$ \\
QP-RoBERTa$_{large}$ & ${45.6}$ & ${43.5}$ & ${56.0}$ \\

 \hline
 \end{tabular}
 }
 \caption{Experiment results on the proposed dataset. 
 (*) This model is the one used to build the dataset. 
QP- models are state-of-the-art cross encoders~\citep{reimers-2019-sentence-bert}.}
 \label{tab:experimental_results}
 \end{table}

\subsection{Comparison with Web-based QA}

We compare QUADRo against Web-based QA systems (WebQA). 
The latter consist of a search engine, BING, which finds a set of documents relevant to the input question, and a state-of-the-art sentence selector~\cite{lauriola2021answer}, which chooses the most probably correct answer sentence, extracted from the retrieved documents. 
This is a typical web-based QA pipeline, the main difference with existing work is that we used BING, a complete and complex commercial search engine, instead of standard solutions based on BM25 or DPR~\cite{10.1145/3511808.3557678,zhang2022situ}.

Our QUADRo configuration consists in S-RoBERTa-QQA used as the retriever and Electra-base-QAQ as the reranker.  
For both systems, QUADRo and WebQA, the answer selectors are applied on top of the top K candidate sentences from the search engines. After preliminary experiments, we set K = 500. The retrieval corpus for QUADRo is set to $6.3$M of q/a pairs, while, the WebQA corpus is defined as the retrieval capability of BING.
To remove biases, we evaluated them with three new sets of questions from open domain sources: Quora, Natural Questions, and TriviaQA, of size 200, 200, and 150, respectively. These questions are not in our DB or dataset. We manually annotated the correctness of the returned answers. 
The results are  reported in Table~\ref{tab:end_to_end_evaluation}. We note that:
First, QUADRo highly outperforms WebQA on Quora 58\% vs 35\%. The main reason is that finding relevant documents for open domain question is always challenging, especially for questions such as Quora where the context is typically underspecified. In contrast, QUADRo can find question that have similar shape and meaning, whose answer is still valid for the target question.

Second, the rationale above is confirmed by the better performance of WebQA on NQ. In this case, the questions that we used have always a relevant Wikipedia page by construction of the NQ dataset. BING is enough powerful to find these pages, so that the only real challenge is the answer selection step. QUADRo still produces a good accuracy, i.e., 50\%.
Finally, QUADRo outperforms WebQA also on TriviQA questions. The latter are rather challenging questions, which are difficult to be found on the web. QUADRo seems to be able to generalize its DB content enough well when operate the search step, or at least better than what a standard search engine can do over web documents.

\begin{table}[t]
\small
\begin{center}

\begin{tabular}{lccc}
    \hline
     \textbf{Model} & \textbf{Quora} & \textbf{NQ} & \textbf{TriviaQA} \\
    \hline
        WebQA  & 35.0 & \textbf{56.0} & 27.0 \\
        QUADRo & \textbf{58.0} & 50.5 & \textbf{29.3} \\
        \hline
        - our dataset & 53.0 & 47.0 & 28.0\\
        - neural SE & 39.0 & 37.5 & \textbf{29.3} \\
        - sentence sel. & 51.5 & 40.0 & 19.0 \\
        - answer relev. & 50.0 & 45.0 & 25.3 \\
    \hline    
\end{tabular}
\end{center}
\caption{End-to-end accuracy of QUADRo and other baselines in 3 open domain QA settings. Best results are highlighted in bold characters.}
\label{tab:end_to_end_evaluation}
\end{table}

%

\subsection{Ablated end-to-end comparisons} 
For completeness, we ran an ablation study to evaluate the impact of each component of our system. The last four rows of 
Table~\ref{tab:end_to_end_evaluation} shows the end-to-end open-domain QA accuracy of QUADRo: 
\vspace{-.3em}
\begin{itemize}
    \item (i) without fine-tuning on our dataset, emphasizing the importance of the resource;\vspace{-.3em}
    \item (ii) when substituting the neural search engine with a standard BM25;\vspace{-.3em}
    \item (iii) without the answer selector (neural search engine only);\vspace{-.3em}
    \item and (iv) with neural SE and selector trained on standard question/question similarity, without using the answer.
\end{itemize}
\vspace{-.3em}
Similarly to the previous comparison, the sentence selectors were applied on top 500 sentences. 

The results show that each element is extremely relevant, and the ensembling of various key technologies make QUADRo an effective QA system.
For example, training on our dataset increases the accuracy by 5\% on Quora, 3.5\% on NQ, and 1.3\% on TriviaQA. The latter small improvement is probably due to the fact that the questions of TiviaQA are rather specific, thus training on general data would impact less than on other datasets.

The usage of neural search engine is essential, our QUADRo using BM25 performs 19\% absolute less than when using our neural ranker. This is rather intuitive as the major advantage of using embeddings is their ability to generalize short text, such as the one constituting questions, while TF$\times$IDF heuristics based on lexical features of BM25 largely suffers from sparse representations.

The answer selector component provides as expected a good contribution, 7-10\% absolute for all datasets.
Finally, the usage of the answer representation as context is rather impactful, from 4\% to 8\%. This demonstrates that, when the target of the evaluation is answer correctness instead of question equivalence, models that take it into account answer context are clearly superior.






\section{Conclusion}


In this paper, we have described our study to scale QA-based on q/a DB to open domain applications.
This required to build a large DB, which we built only using publicly available q/a pairs, reaching a significant size of $\approx 6.3$M items.
To enable retrieval from these large DBs, inspired by the latest neural IR technology, we modeled neural retrieval for q/a pairs. We proposed two different methods based on only questions, and on q/a pairs, where questions and answers can be seen as context. We analyzed the significant impact of using architectures with separate encoder versus dual encoders, in accuracy and efficiency.

This analysis was possible thanks to the design of an accurate dataset for q/a retrieval and ranking, which enables experimentation of the above-mentioned technology. In particular, training and testing ranker models is possible thanks to our design of ranking data.

Finally, we demonstrated that our QA is competitive with strong Web-based methods, built on top the BING search engine. Among other things our work highlights the higher complexity of retrieving and semantically matching questions with answers with respect to retrieving semantically equivalent questions.
Our  data and models, made available to research community, enable interesting future research. For example, how to improve selection of correct q/a pairs in the top $k$ pairs.
\section{Limitations}

The most glaring limit of QUADRo is that the possibility of answering a question strictly depends on the coverage of the database of question/answer pairs. Although the database can be enlarged, covering even more questions, there is still a conceptual limit that prevents the system for answering very infrequent questions. 
Moreover, the database requires mechanisms for content refresh as some questions might change the answer over time.

Concerning the retrieval model, a largely known issue of Dense Retrieval system regards possible drop in performance subject to data domain shift~\cite{wang2021gpl}. 
Although (i) we train the models on various pre-training tasks and open-domain questions, and (ii) our end-to-end experiments shows competitive performance with new data, we cannot quantify this possible issue.

Finally, we observed that, despite the possibility of reading an answer, annotators tend to focus more on the query-question similarity and less on the query-answer relevance. A possible consequence is that models trained on triplets instead of query-question pairs may experience a degradation in performance due to skewed labels. 
Notwithstanding this observation, models trained on query-question pairs work poorly in end-to-end QA evaluation (see Table~\ref{tab:end_to_end_evaluation}).

\bibliography{custom}
\bibliographystyle{acl_natbib}

\appendix
\section{Pre-training}
\label{appendix:pretraining}
Starting from a public checkpoint of our search engine based on Sentence-RoBERTa-base, we continuously pre-trained it on a plethora of datasets for unsupervised STS tasks (paraphrasing, sentence similarity, question answering, and summarization\dots). These datasets include MSMARCO~\cite{nguyen2016ms}, Natural Questions, The Semantic Scholar Open Research Corpus~\cite{lo2020s2orc}, PAQ~\cite{lewis2021paq}, NLI~\cite{snli:emnlp2015}, Altex~\cite{hidey-mckeown-2016-identifying}, AmazonQA~\cite{https://doi.org/10.48550/arxiv.1908.04364}, CNN Dailymail~\cite{see-etal-2017-get}, Coco Captions~\cite{https://doi.org/10.48550/arxiv.1405.0312}, CodeSearchNet~\cite{husain2019codesearchnet}, Eli5~\cite{fan2019eli5}, Fever~\cite{thorne-etal-2018-fever}, Flickr30K~\cite{young-etal-2014-image}, GooAQ, Sentence Compression~\cite{filippova-altun-2013-overcoming}, SimpleWiki~\cite{coster-kauchak-2011-simple}, Specter~\cite{specter2020cohan}, SQuaD~\cite{2016arXiv160605250R}, StackExchange~\cite{Narayan2018DontGM}, WikiHow~\cite{https://doi.org/10.48550/arxiv.1810.09305}, and Xsum~\cite{Narayan2018DontGM}. 

These datasets consist of pairs of semantically equivalent texts (e.g.: question and answer, title and abstract of a document, paraphrasing\dots). 
Overall, the pre-training data includes $\approx180$M positive (i.e. semantically related) text pairs and $\approx17.5$M existing hard-negatives\footnote{Hard negatives are provided for a small subset of pre-training datasets}.
We consider a simple pretraining task where the model predicts if two texts are semantically equivalent or not.
We used the MultipleNegativeRanking~\cite{henderson2017efficient} loss on top of the bi-encoder model combined with cosine similarity as distance metric in order to make the model able to learn powerful embeddings for retrieval. We used a batch-size of 384 and a max sequence length of 256 tokens. We use the AdamW optimizer with a learning rate of $2e^{-5}$.

\section{Reproducibility details}
\label{appendix:retrieval_and_ranking_model}

We used the same strategy to fine tune all models in the experiments showed in this paper.
We used the development split of our proposed resource to select the optimal hyper-parameters configuration through a grid search and to early stop the training when observing a degradation of the validation loss for 2 consecutive epochs. 

The hyper-parameters evaluated for the retrieval (Sentence-RoBERTa) are: the learning rate $\{5e-6, 1e-5, 2e-5, 5e-5\}$, the batch-size $\{64, 128, 256, 384\}$, and the max sequence length $\{128, 256\}$. 
Concerning ranking models (Electra), we tuned the learning-rate $\{5e-6, 1e-5, 3e-5, 5e-5\}$ and the batch-size $\{64, 128, 1024\}$ the max sequence length is set to 256 tokens.

The selection of the initial checkpoint and architecture for the retrieval (RoBERTa) and the reranking (Electra) components was driven by a set of preliminary experiments, where we evaluated multiple Transformer models.
We observed, and reported on table \ref{tab:electra_vs_roberta_results}, that Electra performs better than RoBERTa ($+3.3$ P@1, $+1.9$ MAP, and $+2.5$ MRR.) as cross-encoder in the ranking task measured on our test set. We consider the QAQ configuration as is it the one with better performances. 
Both the models have the same pre-traning as described in the previous sections. 

Differently, Sentence-RoBERTa showed better retrieval performance comapred to Sentence-Deberta-V3 \cite{debertav3} and Sentence-Electra on
measured on STSbenchmark \cite{Cer_2017}, an estabilished benchmark for these tasks. 

\begin{table}[t]
\small
 \centering
 \begin{tabular}{lccc}
\hline
\textbf{{Model}} & \textbf{{P@1}} & \textbf{{MAP}} & \textbf{{MRR}}\\
\hline
{RoBERTa$_{QAQ}$} & {$47.5_{\pm0.3}$} &{$46.5_{\pm0.2}$} & {$57.7_{\pm0.1}$}\\
{Electra$_{QAQ}$} & {$\mathbf{50.8_{\pm0.2}}$} &  {$\mathbf{48.4_{\pm0.1}}$} & {$\mathbf{60.2_{\pm0.1}}$} \\
\hline
 \end{tabular}
 \caption{{Electra and RoBERTa results comparison on the question/answer reranking dataset. The suffix ${QAQ}$ indicates the input setting. }
 }
 \label{tab:electra_vs_roberta_results}
 \end{table}

\begin{table}[t]
\small
 \centering
 \begin{tabular}{lccl}
\hline
\textbf{{Model}} & \textbf{{Pearson corr.}} & \textbf{{Spearman corr.}} \\
 \hline
 {S-RoBERTa} & 
 {$85.4$} & 
{$85.1$} & \\
 
{S-Electra} & 
{$74.8$} & 
{$75.1$} & \\
 
{S-Deberta-V3} & 
{$76.4$} & 
{$77.3$} & \\

\hline
 \end{tabular}
 \caption{{S-RoBERTa vs S-Electra and S-Deberta-V3 performances comparison on the STSbenchmark dataset. All models are base architectures.}
 }
 \label{tab:electra_vs_roberta_retrieval_results}
 \end{table}

\section{Annotators training and guidelines}
\label{appendinx:annotation}

\begin{table*}[t]
\small

\begin{center}
\begin{tabular}{p{.46\linewidth}|p{.46\linewidth}}
    \hline
    \textbf{Positive Examples} & \textbf{Negative Examples}\\
    \hline
    \textbf{Query}: Can a cat and a dog get along?
    
    \textbf{Question}: Do cats like the company of dogs and in the other way around?
    
    \textbf{Answer}: If you are lucky, your cat and dog can become friends within a couple hours. But that won't usually happen. It takes time for cats to adapt to the dogs and similarly for the dogs to learn how to behave around cats.
    
    \textbf{Explanation}: \textit{These questions are both asking if Cats and dogs can be friends. The Answer for the Question is also correct for the Query} 
    & 
    \textbf{Query}: Who did kill Brutus?
    
    \textbf{Question}: Who did Brutus kill?
    
    \textbf{Answer}: Brutus was one of the leaders of the conspiracy that assassinated Julius Caesar
    
    \textbf{Explanation}: \textit{These questions are not asking for the same thing. Moreover the Answer for the Question is not correct for the Query}\\
    \hline 
    \textbf{Query}: Can a person fall in love with another person while he/she is already in love?
    
    \textbf{Question}: Is it possible for people to love 2 person at the same time?
    
    \textbf{Answer}: It is possible to love and be intimate with more than one person at a time.
    
    \textbf{Explanation}: \textit{These questions are both asking if loving 2 people at the same time is possible. The Answer for the Question is correct for both  Question and Query}.
    &
    \textbf{Query}: What is the best restaurant in LA ?
    
    \textbf{Question}: What is the best dish of the best restaurant in LA ?
    
    \textbf{Answer}:The best dish of the best restaurant in LA is Lobster Rolls
    
    \textbf{Explanation}: \textit{Those questions are not asking for the same thing.
    Query asks for a restaurant while the Question asks for a dish.
    Moreover the Answer is not correct for the Query}\\
    \hline
\end{tabular}
\end{center}
\caption{Explained examples used during annotators training.}
\label{tab:training_samples}
\end{table*}

Annotators were asked if two input questions are equivalent or not. The definition of equivalence is: two questions are equivalent iff they have the same intent/meaning and share the same answers. A possible answer for the second question was provided to help the judgment.

We provided a set of guidelines with detailed explained examples to train the annotators. Guidelines consist of a clear description of the task and a set of positive and negative query-question-answer triplets. 
The examples are meant to clarify when query-question pairs can be considered duplicated and how the answer can be used to help the judgment. Some examples used to train the annotators are reported in Table~\ref{tab:training_samples}.

Alongside the guidelines, we introduced a set of control triplets in order to guarantee high annotation quality. Control triplets are designed to be clear, simple and easy to be judged as positive or negative. Table~\ref{tab:control_questions} reports a subset of control triplets used during the annotation. 

Annotators were rewarded with 0.15\$ per annotation task.

\begin{table}[t]
\small
\centering
\begin{tabular}{p{7cm}}
\hline
\textbf{Query:} What is the color of the sun?\\
\textbf{Question:} Which color the sun has?\\
\textbf{Answer:} The sun has a temperature of 5800 Kelvin, so it appears white\\ 
\textbf{Label:} positive \\
\hline
\textbf{Query:} What is the food of Koalas?\\
\textbf{Question:} What do Koalas eat?\\
\textbf{Answer:} Eucalyptus\\
\textbf{Label:} positive \\
\hline
\textbf{Query:} What is the coldest place in the world?\\
\textbf{Question:} What shape is a watermelon?\\
\textbf{Answer:} The watermelons are round or oval shaped\\ 
\textbf{Label:} negative \\
\hline
\textbf{Query:} How many humans are there in the world?\\
\textbf{Question:} What is the color of the strawberry?\\
\textbf{Answer:} Typically they are red\\
\textbf{Label:} negative \\
\hline
\end{tabular}
\caption{Examples of control triplets used to discard annotations.}
\label{tab:control_questions}
\end{table}

\section{Anecdotes}
\label{appendix:anecdotes}

\begin{table}[bt]
\small

\begin{center}
\begin{tabular}{p{1cm}|p{6cm}}
    \textbf{Case \#} & \textbf{Examples}\\
    \hline
    \multirow{4}{*}{\textbf{Case A}} & \textbf{Query}: how old are oldest fossils of organisms?\\
    & \textbf{Question}: The oldest fossils date to how long ago?\\
    & \textbf{Answer}: The oldest documented fossil organisms date to roughly 3.8 billion years ago, shortly after the period of heavy asteroid bombardment in Earth's history.\\
    & \textbf{Label}: 1\\
    \hline
    \multirow{4}{*}{\textbf{Case B}} & \textbf{Query}: where did the term bully pulpit come from\\
    & \textbf{Question}: Who or what is meant by the expression Bully Pulpit?\\
    & \textbf{Answer}: This phrase, used by Theodore Roosevelt, refers to the office of presidency. A pulpit is a place to preach from and "bully" is an older adjective meaning excellent. Therefore a bully pulpit is a great place to make speeches from . Roosevelt appreciated the fact that people listened to him when he spoke as the President.\\
    & \textbf{Label}: 1\\
    \hline
    \multirow{4}{*}{\textbf{Case C}} & \textbf{Query}: What is the name called of a back end of boat?\\
    & \textbf{Question}: What is the name for the back and front of a boat?\\
    & \textbf{Answer}: The front of a vessel be it a ship or boat, is the bow or stem. The back is the stern.\\
    & \textbf{Label}: 0\\
    \hline
    \multirow{4}{*}{\textbf{Case D}} & \textbf{Query}: how how to replace battery in liftmaster remote control?\\
    & \textbf{Question}: how to change battery in liftmaster remote keypad?\\
    & \textbf{Answer}: 'Slide the battery cover down. The battery is located at the bottom of your keypad.', 'Disconnect the old battery and remove it. Remember to dispose of your old battery correctly.', 'Install the new battery. ... ', 'Put the battery cover back in place.'\\
    & \textbf{Label}: 1\\

    \hline
\end{tabular}
\end{center}
\caption{\label{tab:anecdotes_samples}Annotated triplets from our dataset.}
\end{table}

This section shows and discusses some examples of produced annotations, showed in Table~\ref{tab:anecdotes_samples}.
In the first example, Case A, query and question are closer in terms of question wording and are asking for the same thing make it easy to label them as duplicates even without taking the answer into account. By contrast, in Case B we can notice that query and question seem to be asking for different things. In this case, the answer played a key role for the final annotation label, since it answers to both the query and the question. The assigned positive label is then correct.

Case C contains an annotation error. In this case, the query is contained in the question. While the question is asking for the name of the back and the front of a boat, the query is asking only for the name of the back of a boat. Without considering the answer, an annotator might be prone to label these questions as non-duplicated, since they are asking for slightly different concepts. 
However, in a end-to-end QA setting, the answer is correct for the input query (and the question), and thus we should consider the triplet as positive.

In the last example, Case D, we report a wrong positive annotation. The query and the questions seems to be identical, but they are not. The query is asking how to replace the battery from a liftmaster remote control, while the question is asking how to replace the battery of a liftmaster remote keypad. In this case the query and the question are referring to two distinct devices. Moreover the provided answer is not correct with respect to the query. In this case, the correct annotation should be negative.

\section{Latency}
\label{appendix:latency}


We conducted a latency analysis on QUADRo, evaluating the efficiency with respect to various key aspects, including the number of retrieved q/a pairs and the size of the database. 
In our tests, we used (i) a Nvidia A100 GPU, (ii) a S-RoBERTa (base) retrieval model with embedding size of 768, and (iii) an Electra-base reranker.

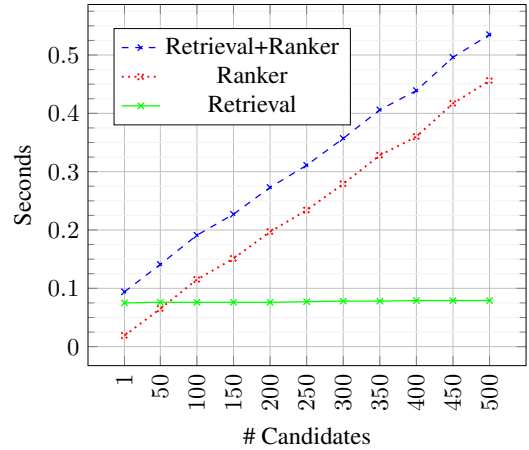
\begin{figure}
\resizebox{0.9\linewidth}{!}{
\begin{tikzpicture}
\begin{axis}[
	xlabel=\# Candidates,
	ylabel= Seconds,
    legend style={at={(.6,.8)},anchor=east},
    xtick={1,50,100,150,200,250,300,350,400,450,500},
    ytick={0,0.1,0.2,0.3,0.4,0.5,0.6,0.7,0.8,0.9},
    grid=both,
    grid style={line width=.1pt, draw=gray!10},
    major grid style={line width=.2pt,draw=gray!50},
    minor grid style={line width=.2pt,draw=gray!10},
    minor tick num=3,
    x tick label style={rotate=90,anchor=east}
    ]



\addplot[dashed,color=blue,mark=x,line width=.2mm] coordinates {
	(1, 0.094)
    (50, 0.141)
    (100, 0.191)
    (150, 0.227)
    (200, 0.273)
    (250, 0.311)
    (300, 0.357)
    (350, 0.406)
    (400, 0.439)
    (450, 0.496)
    (500, 0.535)
};

\addplot[dotted, color=red,mark=x,line width=.3mm] coordinates {
	(1, 0.019)
    (50, 0.065)
    (100, 0.115)
    (150, 0.151)
    (200, 0.197)
    (250, 0.234)
    (300, 0.279)
    (350, 0.328)
    (400, 0.360)
    (450, 0.417)
    (500, 0.456)
};

\addplot[solid, color=green,mark=x,line width=.2mm] coordinates {
	(1, 0.075)
    (50, 0.076)
    (100, 0.076)
    (150, 0.076)
    (200, 0.076)
    (250, 0.077)
    (300, 0.078)
    (350, 0.078)
    (400, 0.079)
    (450, 0.079)
    (500, 0.079)
};

\legend{Retrieval+Ranker, Ranker, Retrieval}
\end{axis}

\end{tikzpicture}}
\caption{Latency of the end-to-end QUADRo system while increasing the number of retrieved and reranked q/a pairs. Values are averaged over 200 executions.}
\label{fig:latency_candidates}
\end{figure}

\figurename~\ref{fig:latency_candidates} show the latency of an end-to-end request when varying the number of the retrieved q/a pairs while using a database of $\approx6.3$M q/a pairs. 
As you can see, the time scales linearly with the amount of retrieved data. To retrieve and rerank 500 q/a pairs QUADRo took only $\approx0.53$s, 140ms to retrieve and rerank 50 pairs. According to \figurename~\ref{fig:latency_candidates} we can notice that the majority of the time is spent in the ranking process rather than in retrieving the pairs from the database. The retrieval time is $\approx 77$ms, and it does not scale as the number of returned q/a pairs increases. 

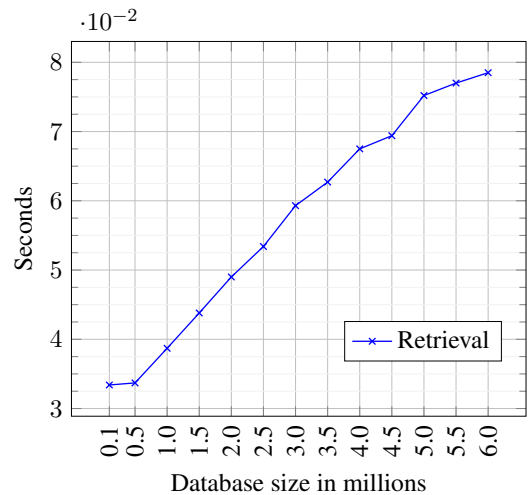
\begin{figure}
\resizebox{0.9\linewidth}{!}{
\begin{tikzpicture}
\begin{axis}[
	xlabel= Database size in millions,
	ylabel= Seconds,
    legend style={at={(.6,.2)},anchor=west},
    xtick={0.1,0.5,1.0,1.5,2.0,2.5,3.0,3.5,4.0,4.5,5.0,5.5,6.0},
    xticklabels={0.1,0.5,1.0,1.5,2.0,2.5,3.0,3.5,4.0,4.5,5.0,5.5,6.0},
    grid=both,
    grid style={line width=.1pt, draw=gray!10},
    major grid style={line width=.2pt,draw=gray!50},
    minor grid style={line width=.2pt,draw=gray!10},
    minor tick num=3,
    x tick label style={rotate=90,anchor=east}
    ]

\addplot[color=blue,mark=x,line width=.2mm] coordinates {
	(0.1, 0.0334)
    (0.5, 0.0337)
    (1.0, 0.0387)
    (1.5, 0.0438)
    (2.0, 0.0490)
    (2.5, 0.0534)
    (3.0, 0.0593)
    (3.5, 0.0627)
    (4.0, 0.0675)
    (4.5, 0.0694)
    (5.0, 0.0752)
    (5.5, 0.0770)
    (6.0, 0.0785)
};

\legend{Retrieval}
\end{axis}

\end{tikzpicture}}
\caption{Latency of the end-to-end QUADRo system while increasing the dimension of the DB. Values are averaged over 200 executions.}
\label{fig:latency_database_size}
\end{figure}

\figurename~\ref{fig:latency_database_size} shows the latency of the system while increasing the dimension of the DB. In the experiment we set the number of retrieved q/a pairs to 500. The retrieval time scales as the size of the database increases. Is worth to mention that the retrieval process is in the magnitude order of milliseconds, confirming the efficiency of our system. 

We release the implementation and the code to replicate these experiments upon requests. 
The experiments were based on PyTorch 1.9, Cuda 10.2, Python 3.8, and Transformers (by HuggingFace) 4.10.

\end{document}